%% file: main.tex
\documentclass{article}

\usepackage[final]{neurips_2022}

\usepackage[utf8]{inputenc} 
\usepackage[T1]{fontenc}    
\usepackage{hyperref}       
\usepackage{url}            
\usepackage{booktabs}       
\usepackage{amsfonts}       
\usepackage{nicefrac}       
\usepackage{microtype}      
\usepackage{xcolor}         
\usepackage{amsmath}
\usepackage{amssymb}
\usepackage{graphicx}
\usepackage{mathtools}
\usepackage{mymacros}

\title{Remote Sensing-Based Assessment of Economic Development}

\author{Yijian Pan, Yongchang Ma, Bolin Shen, Linyang He}

\begin{document}

\maketitle

\section{Introduction}
The goal of our project is to use satellite data (including nighttime light data and remote sensing images) to give us some statistical estimation of the economic development level of a selected area (Singapore). Findings from the project could inform policymakers about areas needing intervention or support for economic development initiatives. Insights gained might aid in targeted policy formulation for infrastructure, agriculture, urban planning, or resource management.

\section{Economic Activity Estimation Based on Light Intensity}
In this section, we focus on estimating economic activity based on light intensity from nighttime light (NLT) data. However, the raw nighttime light data suffers from the problems of interannual inconsistency, saturation, and blooming. Researchers have proposed several methods to address these issues. Unfortunately, their methods only overcome one or more of these problems and a global data product for all the solvable issues has yet to emerge. Therefore, we aim to produce a consistent and corrected NLT dataset before we predict GDP based on the light intensity from NLT data.
\subsection{Interannual inconsistency correction}
\input{intercalibration}
\subsection{Saturation correction}
\input{saturation}
\subsection{Blooming efect correction}
\input{blooming}
\subsection{Prediction based on linear regression models}
\input{pred_reg}

\section{Economic Activity Estimation Based on Boat Intensity}
The second part of this project uses computer vision technology to automatically count the number of ships in Singapore ports, and compares and analyzes these data with official port throughput and GDP data.
This section emphasizes the important position of Singapore's ports in international trade and the importance of accurate logistics data for economic analysis. This helps governments and businesses better understand and predict economic trends. For this part, we consider the Port of Singapore. 
\subsection{Sentinel-2}
The satellite images for this project come from Sentinel-2, which is part of the European Space Agency (ESA) and is a satellite mission dedicated to Earth observation. The satellite is primarily used for environmental monitoring and land management, providing high-resolution multispectral imaging. Sentinel-2 can cover the entire globe, including land, coastlines and inland waters. Sentinel-2 has a revisit period of about five days, meaning it can capture images of the same spot on Earth at least once every five days.
\subsection{Boats detection in the Singapore Port}
To calculate the number of ships in Singapore ports, we use the following pipeline:

\textbf{(1)Satellite Data acquisition}: First determine the specific coordinate area of the Singapore port to ensure that correct satellite image data is obtained. Use the Python API interface provided by Sentinel-2 to obtain satellite image data of the same location from 2017 to 2019. These data include continuous observation records in the Singapore port area.

\textbf{(2)Time and weather filtering}: Since the time accuracy of Sentinel-2 is up to 5 days and is affected by weather, we eliminated unrecognizable images, such as those covered by clouds.

\textbf{(3)Image segmentation}: In order to improve the detection efficiency and accuracy of the subsequent computer vision model, the image of the Singapore port area was divided into 2×6 and a total of 12 blocks. Each segmented image is input into the computer vision model for ship detection. Subsequently, the detection results of these blocks are aggregated to obtain the total number of ships in the entire port area.

\begin{figure}[h]
\centering
\includegraphics[width=0.9\textwidth]{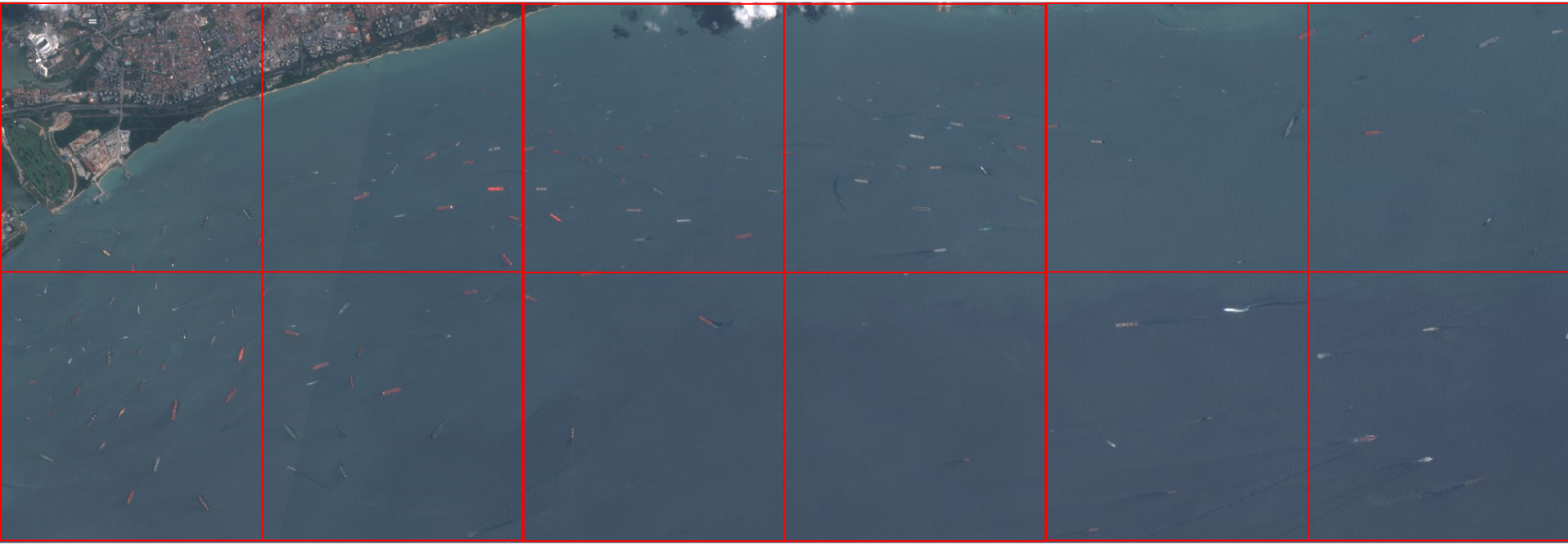}
\caption{Singapore port's boats}
\end{figure}

\subsection{YOLO}
We tried YOLOv5, YOLOv8, and DETR for our boat detection task. DETR is an end-to-end object detection network using Transformers. We tried fine-tuning this network, but it turned out that the time was unacceptable given we don't have any GPU. So at last we used YOLOv8x to finish this task. YOLOv8 is a cutting-edge, state-of-the-art (SOTA) model that builds upon the success of previous YOLO versions and introduces new features and improvements to further boost performance and flexibility. The detailed architecture and description of YOLO models can be seen here. (You Only Look Once: Unified, Real-Time Object Detection)

Compared to the model, the dataset plays a much more important role. No dataset is based on Sentinel-2 satellite data, but we do find a dataset that uses Google Earth satellite images to detect boats near harbors. The dataset is in https://www.kaggle.com/datasets/tomluther/ships-in-google-earth., it contains 794 jpegs showing various-sized ships in Google Earth satellite imagery. It's similar to our needs, so we chose this dataset as the foundation.

The images from different satellites can differ a lot, because different satellites have different resolutions, exposure, contrast, saturation, etc. Although they seem similar, the domain gap is not neglectable. We don't want to create a dataset based on Sentinel-2 data, so we cannot use many classical domain adaptation methods. We can only do something on this existing dataset and try to improve the performance of our model in Sentinel-2 data.

The first thing we do is to resize all the images to the fixed size: 640*640 pixels. YOLO is trained on this size of images, so it is better to use the same size of images. To maintain the aspect ratio, we fill the rest of the image with black pixels.

The dataset only contains 794 images, which is quite small. Therefore, we use some common data augmentation methods like rotation, crop, flip, etc.

Different satellites have different exposure and contrast. To align different satellites, we use some extra data augmentation methods that randomly increase/decrease the contrast/exposure/brightness of the image. After applying these data augmentation methods, the detection map increases a lot.

Finally, to align different resolutions of Google Earth satellites and Sentinel-2, we use nested mosaic data augmentation. We found that the resolution of our data is much worse than Google Earth, so the performance of our model in Sentinel-2 is much worse than in Google Earth. To alleviate this phenomenon, we combined different images together in a nested manner. This means some augmented images contain several raw images with different sizes. After doing this, the size of boats varies a lot, and there are many small boats in the dataset with similar size to that in Sentinel-2.

After doing all these augmentations, our model performs much better than just using the raw dataset.

The training process and network are not modified, and we use the preset of YOLOv8 in ultralytics.

\begin{figure}[h]
\centering
\includegraphics[width=0.35\textwidth]{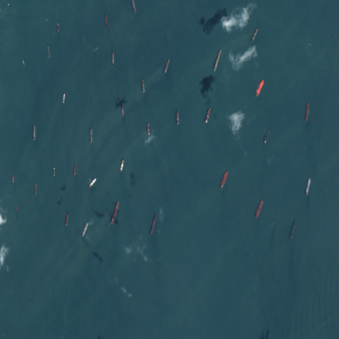}
\includegraphics[width=0.35\textwidth]{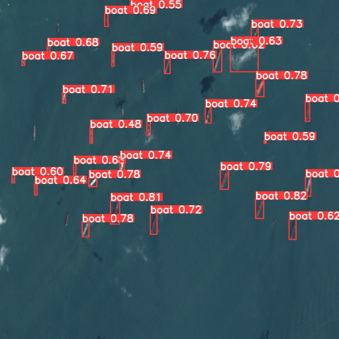}
\caption{Boat detection results using YOLO (left:input, right:output)}
\end{figure}

\section{Results}
\subsection{GDP prediction based on nighttime light data}
\input{experiment1}

\subsection{Number of boats and GDP, port throughput}
For the second part of our project, we compare the annual boat count obtained using the above method with the GDP data and port throughput released by the Singapore government. We found that the number of ships showed a significant growth trend from 2017 to 2018, while from 2018 to 2019 it tended to be flat with little change. Naturally, this coincides with the port throughput data and, more interestingly, with the GDP data. This suggests that in addition to the brightness of city lights, the number of ships in the port is also an economic indicator that can be referenced.

\begin{figure}[h!]
\centering
\includegraphics[width=0.6\textwidth]{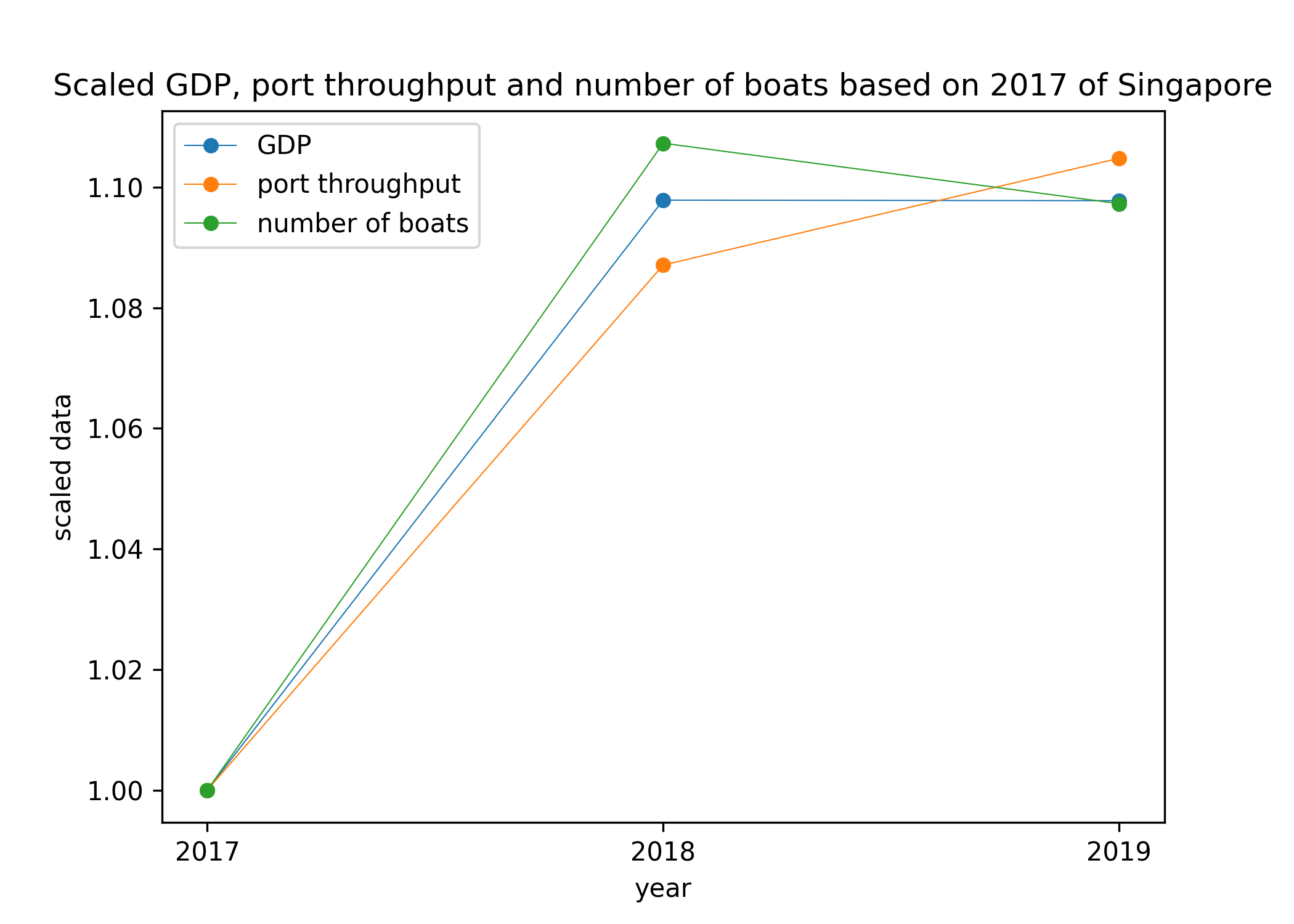}
\caption{Singapore port's boat number, port throughput and GDP (scaled based on 2017)}
\end{figure}

\section{Code Availability}
The source code for the project is available at \href{https://github.com/Edlison/EECS504-Project}{https://github.com/Edlison/EECS504-Project}.
The file, nighttime light data processing.txt, needs to be run in the Earth Engine Code Editor.

\newpage
\section{References}
{
\small
[1]Wu, Jiansheng, et al. “Intercalibration of DMSP-OLS Night-Time Light Data by the Invariant Region Method.” International Journal of Remote Sensing, vol. 34, no. 20, 2013, pp. 7356–7368.

[2]Hu, Yang, and Jin Chen. “Correcting the Saturation Effect in DMSP/OLS Stable Nighttime Light Products Based on Radiance-Calibrated Data | IEEE Journals and Magazine | IEEE Xplore.” Ieeexplore.ieee.org, 1 Mar. 2021, ieeexplore.ieee.org/document/9366289.

[3]Cao, Xin, et al. “A Simple Self-Adjusting Model for Correcting the Blooming Effects in DMSP-OLS Nighttime Light Images.” Remote Sensing of Environment, vol. 224, 1 Apr. 2019, pp. 401–411.

[4]Doll, Christopher N.H., et al. “Mapping Regional Economic Activity from Night-Time Light Satellite Imagery.” Ecological Economics, vol. 57, no. 1, 2006, pp. 75–92.

[5]Bennett, Mia M., and Laurence C. Smith. “Advances in Using Multitemporal Night-Time Lights Satellite Imagery to Detect, Estimate, and Monitor Socioeconomic Dynamics.” Remote Sensing of Environment, vol. 192, Apr. 2017, pp. 176–197.

}

\section{Appendix}
\begin{figure}[h!]
\centering
\includegraphics[width=0.9\textwidth]{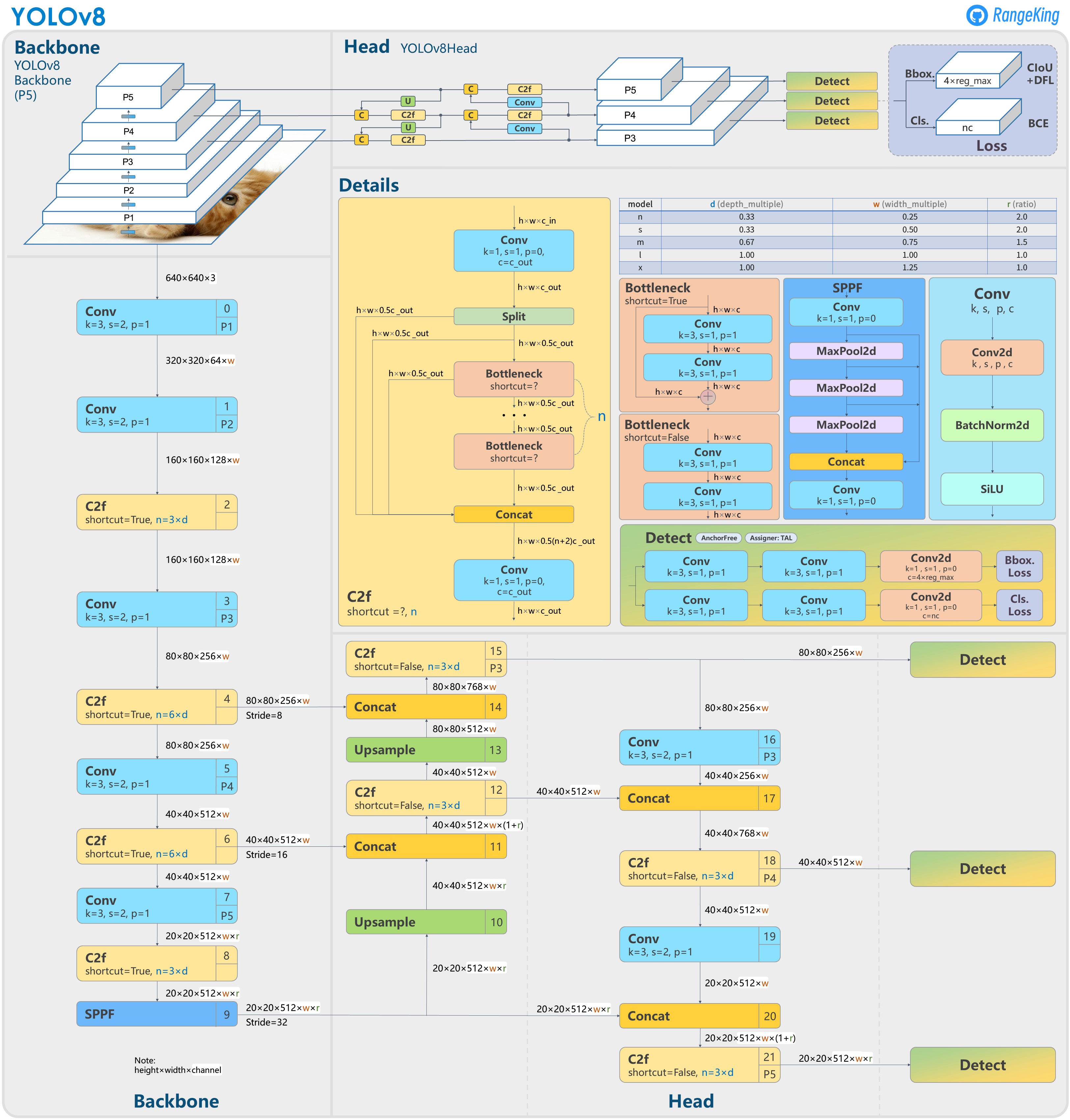}
\caption{YOLOv8 Architecture}
\end{figure}

\end{document}

%% file: intercalibration.tex
The Earth is approximately spherical, while remote sensing images are flat representations. This difference causes an apparent interannual inconsistency in the sum of NTL digital number (DN) values at global and national scales. To correct the inconsistency, we need to make use of the invariant features in the images. Mauritius, Puerto Rico, and Okinawa are selected as invariant regions. We used the selected invariant regions as the mask, extracting the invariant area of the pending images and reference image, reading the greyscale matrices of each region, and depositing them in arrays. We then performed a regression analysis of the arrays corresponding to the pending images and reference image, framed the regression model, and acquired the intercalibration models corresponding to each pending image. The regression model is as follows:
\begin{equation}
    DN_c+1 = a\times(DN_m+1)^b
\end{equation}
where $DN_c$ is the pixel value after correction, $DN_m$ is the original DN value, $a$ and $b$ are the unknown coefficients in the model.

%% file: saturation.tex
Saturation effect occurs when the sensor or detector onboard a satellite collects more photons (or light) than it can accurately measure or record. We need auxiliary data, radiance calibrated nighttime light product, to correct the saturation effect. Radiation calibrated data significantly eliminates the saturation effect by synthesizing data at different gains and has substantial advantages in spatial analysis. However, the number of images is relatively small to meet the requirements of our studies, so we use the images to calibrate NLT data. 
We identifed the region with a DN value of 63 as the saturation region. Sample pixels were selected based on the difference between the NLT data and the radiance-calibrated data from neighboring years. Saturation zone DN values can be obtained from the corresponding area of the radiance-calibrated data and the logarithmic model. The specific equation is as follows:
\begin{equation}
    DN_{LM} = a\times log(DN_R)+b
\end{equation}
where $DN_{LM}$ is the DN value corrected by the logarithmic model, $DN_R$ is the radiance-calibrated data DN value of the saturated zone, $a$ and $b$ are the coefficients of the regression model.

%% file: blooming.tex
Blooming effect is similar to saturation effect, but it causes a spillover of light into neighboring pixels instead of a specific pixel. We assume that a pseudo light pixel (i.e., a bright pixel adjacent to the background) should have no light, and its value is contributed by the blooming effect of other bright pixels around it. The blooming effect can be quantitatively described by a spatial response function with pseudo light pixels as 
samples, and the specific equation is as follows:
\begin{equation}
    R'=a\times \sum_{i=1}^{N} \frac{R_i}{d_i^2} +b
\end{equation}
where $R'$ is the value of brightness change due to the blooming effect, 
$R_i$ represents the pixel value of the neighboring pixels in the moving window, $N$ is the number of neighboring pixels, $d_i$ is the Euclidean distance from the pseudo-pixel, $a$ and $b$ are coefficients describing the blooming effect.

%% file: pred_reg.tex
For this downstream task, we will utilize nighttime light intensity as the primary predictive variable. Light intensity is derived from NLT data which is corrected through the above three steps.

Corresponding GDP data, which serves as the dependent variable, will be aligned with the night light intensity. This GDP data is sourced from reliable economic databases of Singapore and reflects the economic output of Singapore corresponding to the night light intensity.

The data set spans 1992-2011, allowing for a comprehensive analysis of trends over time. We employ a linear regression model to establish the relationship between night light data and GDP. The choice of a linear model is grounded in the initial hypothesis that changes in illumination intensity are linearly correlated with changes in economic activity, as represented by GDP. The linear regression model is formulated as follows:

\[
GDP = \beta_0 +\beta_1 * Night Light + \epsilon
\]

where $\beta_0$ is the intercept, $\beta_1$ is the coefficient for night light data, and $\epsilon$ represents the error term. To evaluate the performance of our linear regression model, we will use the Mean Squared Error (MSE) metric. MSE is calculated as:

\[
MSE = {1\over n}\sum_{i=1}^n(Y_i-\hat{Y}_i)^2
\]

where $Y_i$ is the actual GDP value, $\hat{Y}_i$ is the predicted GDP value by the model, and $n$ is the number of observations. A lower MSE value indicates a better fit of the model to the data.

%% file: experiment1.tex
\begin{figure}[h]
\includegraphics[width=0.5\textwidth]{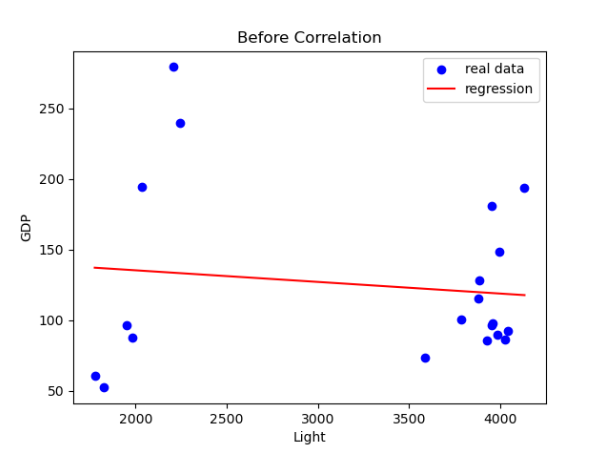}
\includegraphics[width=0.5\textwidth]{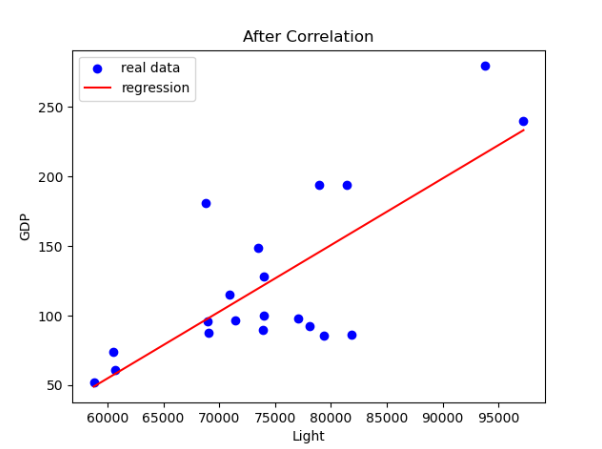}
\caption{Regression result using illumination data before and after correlation}
\end{figure}

This experiment was to use linear regression to predict the GDP for the year 2012 using night light data. The model was trained on night light and corresponding GDP data spanning from 1992 to 2011. The training involved establishing a relationship between the intensity and distribution of night light illumination and the economic output as measured by GDP. Using the model trained on historical data, we predicted the GDP for the year 2012, solely based on the night light data of that year.

The actual reported GDP for the year 2012 was 295.09. When the model used night light data that showed a high correlation with historical GDP data, the predicted GDP for 2012 was 227.37. The Mean Squared Error (MSE) in this case was calculated to be 4586.61. Conversely, using night light data with low or no correlation to GDP resulted in a significantly different prediction. The predicted GDP for 2012 in this scenario was 133.33, with a substantially higher MSE of 26165.73.

These findings suggest that the effectiveness of using night light data to predict GDP hinges on the degree of correlation between night light intensity and economic activity. It implies that more accurate predictions can be made if the illumination data is carefully selected and validated for its economic representativeness.